\documentclass[10pt,twocolumn,letterpaper]{article}

\usepackage{cvpr}
\usepackage{times}
\usepackage{epsfig}
\usepackage{graphicx}
\usepackage{amsmath}
\usepackage{amssymb}
\usepackage{color}
\usepackage{comment}
\usepackage[ruled]{algorithm2e}
\usepackage{wrapfig}
\usepackage{makecell}
\usepackage{multirow}
\usepackage{microtype}

\newcommand{\surrog}[1]{\overline{#1}}
\newcommand{\newpos}[1]{\widehat{#1}}
\newcommand{\nua}{\nu_{\textrm{a}}}
\newcommand{\nur}{\nu_{\textrm{r}}}
\DeclareMathOperator*{\argmin}{arg\,min}
\DeclareMathOperator*{\proj}{proj}
\newcommand{\SOG}{\mathcal{R}}
\DeclareMathOperator{\Tr}{Tr}

\usepackage[pagebackref=true,breaklinks=true,letterpaper=true,colorlinks,bookmarks=false]{hyperref}

\cvprfinalcopy % *** Uncomment this line for the final submission

\def\cvprPaperID{5011} % *** Enter the CVPR Paper ID here

\ifcvprfinal\fi
\begin{document}

%%%%%%%%% TITLE
\title{Quasi-Newton Solver for Robust Non-Rigid Registration}

\author{Yuxin Yao\textsuperscript{1} \quad Bailin Deng\textsuperscript{2} \quad Weiwei Xu\textsuperscript{3} \quad Juyong Zhang\textsuperscript{1}\thanks{Corresponding author}\\
	\textsuperscript{1}University of Science and Technology of China \quad \textsuperscript{2}Cardiff University\quad \textsuperscript{3}Zhejiang University\\
	{\tt\small yaoyuxin@mail.ustc.edu.cn} \quad  {\tt\small DengB3@cardiff.ac.uk}\quad  {\tt\small xww@cad.zju.edu.cn} \quad  {\tt\small juyong@ustc.edu.cn}
%\and
}

\maketitle
%\thispagestyle{empty}

%%%%%%%%% ABSTRACT
\begin{abstract}
Imperfect data (noise, outliers and partial overlap) and high degrees of freedom make non-rigid registration a classical challenging problem in computer vision. Existing methods typically adopt the $\ell_{p}$ type robust estimator to regularize the fitting and smoothness, and the proximal operator is used to solve the resulting non-smooth problem. However, the slow convergence of these algorithms limits its wide applications. In this paper, we propose a formulation for robust non-rigid registration based on a globally smooth robust estimator for data fitting and regularization, which can handle outliers and partial overlaps. We apply the majorization-minimization algorithm to the problem, which reduces each iteration to solving a simple least-squares problem with L-BFGS. Extensive experiments demonstrate the effectiveness of our method for non-rigid alignment between two shapes with outliers and partial overlap, with quantitative evaluation showing that it outperforms state-of-the-art methods in terms of registration accuracy and computational speed. The source code is available at \url{https://github.com/Juyong/Fast_RNRR}.
\end{abstract}

\section{Introduction}

With the popularity of depth acquisition devices such as Kinect, PrimeSense and the depth sensors on smartphones, techniques for 3D object tracking and reconstruction from point clouds have enabled various applications. Non-rigid registration is a fundamental problem for such techniques, especially for the reconstruction of dynamic objects.
Since depth maps obtained from structured light or time-of-flight cameras often contain outliers and holes, a robust non-rigid registration algorithm is needed to handle such data. Moreover, real-time applications require high computational efficiency for non-rigid registration.

Given two point clouds sampled from a source surface and a target surface respectively,
the aim of non-rigid registration is to find a transformation field for the source point cloud to align it with the target point cloud.
This problem is typically solved via optimization.
The objective energy often includes alignment terms that measure the deviation between the two point clouds after the transformation, as well as regularization terms that enforce smoothness of the transformation field.
Existing methods often formulate these terms using the $\ell_2$-norm, which penalizes alignment and smoothness errors across the whole surface~\cite{amberg2007optimal,li2008global,li2009robust}. On the other hand, the ground-truth alignment can potentially induce large errors for these terms in some localized regions of the point clouds, due to noises, outliers, partial overlaps, or articulated motions between the point clouds. Such large localized errors will be inhibited by the $\ell_2$ formulations, which can lead to erroneous alignment results.
To improve the alignment accuracy, recent works have utilized sparsity-promoting norms for these terms, such as the $\ell_1$-norm~\cite{yang2015sparse,li2018robust,jiang2019huber} and the $\ell_0$-norm~\cite{guo2015robust}.
The sparsity optimization enforces small error metrics on most parts of the point cloud while allowing for large errors in some local regions, improving the robustness of the registration process.
However, these sparsity terms can lead to non-smooth problems that are more challenging to solve.
Existing methods often employ first-order solvers such as the alternating direction method of multipliers (ADMM), which can suffer from slow convergence to high-accuracy solutions~\cite{boyd2011distributed}.

In this paper, we propose a new approach for robust non-rigid registration with fast convergence. The key idea is to enforce sparsity using the Welsch's function~\cite{holland1977robust}, which has been utilized for robust processing of images~\cite{ham2015robust} and meshes~\cite{zhang2018static}.
We formulate an optimization that applies the Welsch's function to both the alignment error and the regularization, to achieve robust registration.
Unlike the $\ell_p$-norms, the Welsch's function is smooth and does not induce non-smooth optimization.
We solve the optimization problem using the majorization-minimization (MM) algorithm~\cite{Lange2004}. It iteratively constructs a surrogate function for the target energy based on the current variable values and minimizes the surrogate function to update the variables, and is guaranteed to converge to a local minimum.
The Welsch's function enables us to derive a surrogate function in a simple least-squares form, which can be solved efficiently using L-BFGS. Experimental results verify the robustness of our method as well as its superior performance compared to existing robust registration approaches.

In summary, the main contributions of this paper include:
\begin{itemize}
  \item We formulate an optimization problem for non-rigid registration, using the Welsch's function to induce sparsity for alignment error and transformation smoothness. The proposed formulation effectively improves the robustness and accuracy of the results.
  \item We propose an MM algorithm to solve the optimization problem, using L-BFGS to tackle the sub-problems. The combination of MM and L-BFGS greatly improves the computational efficiency of robust non-rigid registration compared to existing approaches.
\end{itemize} 
\section{Related work}
Non-rigid registration has been widely studied in computer vision and image processing. The reader is referred to~\cite{TamCLLLMMSR13} for a recent survey on rigid and non-rigid registration of 3D point clouds and meshes. In the following, we focus on works that are closely related to our method.

Various optimization-based methods have been proposed for non-rigid registration. Chui et al.~\cite{Chui2000A} utilized a Thin Plate Spline (TPS) model to represent non-rigid mappings, and alternately update the correspondence and TPS parameters to find an optimized alignment. Following this approach, Brown et al.~\cite{brown2007Global} used a weighted variant of iterative closest point (ICP) to obtain sparse correspondences, and warped scans to global optimal position by TPS. Extending the classical ICP algorithm for rigid registration,  Amberg et al.~\cite{amberg2007optimal} proposed a non-rigid ICP algorithm that gradually decreases the stiffness of regularization and incrementally deforms the source model to the target model. Li et al.~\cite{li2008global} adopted an embedded deformation approach~\cite{sumner2007embedded} to express a non-rigid deformation using transformations defined on a deformation graph, and simultaneously optimized the correspondences between source and target scans, confidence weights for measuring the reliability of correspondence, and a warping field that aligns the source with the target. Later, Li et al.~\cite{li2009robust} combined point-to-point and point-to-plane metrics for the more accurate measure of correspondence.

Other methods tackle the problem from a statistical perspective.  Considering the fitting of two point clouds as a probability density estimation problem, Myronenko et al.~\cite{Andriy2010Point} proposed the Coherent Point Drift algorithm which encourages displacement vectors to point into similar directions to improve the coherence of the transformation.
Hontani et al.~\cite{hontani2012robust} incorporated a statistical shape model and a noise model into the non-rigid ICP framework, and detected outliers based on their sparsity. Jian et al.~\cite{jian2005robust} represented each point set as a mixture of Gaussians treated point set registration as a problem of aligning two mixtures. Also with a statistical framework, Wand et al.\cite{Wand2009Efficient} used a meshless deformation model to perform the pairwise alignment. Ma et al.~\cite{ma2015robust} proposed an ${L_2}E$ estimator to build more reliable sparse and dense correspondences.

Many non-rigid registration algorithms are based on $\ell_2$-norm metrics for the deviation between source and target models and the smoothness of transformation fields, which can lead to erroneous results when the ground-truth alignment induces large deviation or non-smooth transformation due to noises, partial overlaps, or articulate motions.
To address this issue, various sparsity-based methods have been proposed. Yang et al.~\cite{yang2015sparse} utilized the $\ell_1$-norm to promote regularity of the transformation. Li et al.~\cite{li2018robust} additionally introduced position sparsity to improve robustness against noises and outliers. For robust motion tracking and surface reconstruction, Guo et al.~\cite{guo2015robust} proposed an $\ell_0$-based motion regularizer to accommodate articulated motions.

Depending on the application, different regularizations for the transformation have been proposed to improve the robustness of the results. In~\cite{dyke2019non}, an as-rigid-as-possible energy was introduced to avoid shrinkage and keep local rigidity. Wu et al.~\cite{wu2019global} introduced an as-conformal-as-possible energy to avoid mesh distortion. Jiang et al.~\cite{jiang2019huber} applied a Huber-norm regularization to induce piecewise smooth transformation. 
\section{Problem Formulation}
Let $\mathcal{S}=\{\mathcal{V},\mathcal{E}\}$ be a source surface consisting of sample points $\mathcal{V}=\{\mathbf{v}_1, \mathbf{v}_2, \ldots, \mathbf{v}_n \in \mathbb{R}^3\}$ connected by a set of edges $\mathcal{E}$. Let $\mathcal{T}$ be a target surface with sample points $\mathcal{U}=\{\mathbf{u}_1, \mathbf{u}_2, \ldots, \mathbf{u}_m \in \mathbb{R}^3 \}$.
We seek to apply affine transformations to the source points $\mathcal{V}$ to align them with the target surface.
In this paper, we adopt the embedded deformation approach proposed in~\cite{sumner2007embedded} to model the transformations. Specifically, we construct a \emph{deformation graph} $\mathcal{G}$ with its vertices $\mathcal{V}_{\mathcal{G}} = \{\mathbf{p}_1,\ldots,\mathbf{p}_r\}$ being a subset of the source points $\mathcal{V}$, and with its edges $\mathcal{E}_{\mathcal{G}}$ connecting vertices that are nearby on the target surface.
On each deformation graph vertex $\mathbf{p}_j$ we define an affine transformation represented with a transformation matrix $\mathbf{A}_j \in \mathbb{R}^{3 \times 3}$ and a displacement vector $\mathbf{t}_j \in \mathbb{R}^3$. Each vertex $\mathbf{p}_j$ influences a localized region that contains any source point $\mathbf{v}_i$ whose geodesic distance $D(\mathbf{v}_i, \mathbf{p}_j)$ to $\mathbf{p}_j$ on the source surface is smaller than a user-specified radius $R$. If $\mathbf{p}_j$ influences a source point $\mathbf{v}_i$, the affine transformation $(\mathbf{A}_j, \mathbf{t}_j)$ associated with $\mathbf{p}_j$ induces a transformed position $\mathbf{A}_j(\mathbf{v}_i-\mathbf{p}_j)+\mathbf{p}_j + \mathbf{t}_j$ for $\mathbf{v}_i$.
Then the final transformed position $\newpos{\mathbf{v}}_i$ for $\mathbf{v}_i$ is a convex combination of all positions induced by the vertices in graph $\mathcal{G}$~\cite{li2009robust}:
\begin{equation}
\label{eq:sample_represent}
\newpos{\mathbf{v}}_i =  \sum_{\mathbf{p}_j \in \mathcal{I}(\mathbf{v}_i)} {w}_{ij} \cdot \left(\mathbf{A}_j(\mathbf{v}_i-\mathbf{p}_j)+\mathbf{p}_j + \mathbf{t}_j\right),
\end{equation}
where $\mathcal{I}(\mathbf{v}_i) = \{\mathbf{p}_j \mid D(\mathbf{v}_i, \mathbf{p}_j)<R \}$ denotes the set of vertices that influence $\mathbf{v}_i$, and ${w}_{ij}$ is a distance-dependent normalized weight:
\[
    {w}_{ij} = \frac{\left( 1 - {D^2(\mathbf{v}_i, \mathbf{p}_j)}/{R^2}\right)^3}
    {\sum_{\mathbf{p}_k \in \mathcal{I}(\mathbf{v}_i)} \left( 1 - {D^2(\mathbf{v}_i, \mathbf{p}_k)}/{R^2}\right)^3}.
\]
Compared to formulations that attach a different transformation to each source point such as~\cite{li2018robust}, a major benefit of using the deformation graph is that the deformation of the target surface can be defined with a much smaller number of variables, enabling more efficient optimization.

In the following, we first present an optimization formulation to determine the affine transformations associated with the deformation graph, and then explain how to construct the deformation graph. For the ease of presentation, we use $\mathbf{X}_j$ to denote the affine transformation $(\mathbf{A}_j, \mathbf{t}_j)$ at vertex $\mathbf{p}_j$, whereas $\mathbf{X}$ denotes the set of all transformations.

\subsection{Optimization Formulation}
\label{seq:model}

For robust alignment between the source and the target surfaces, we determine the affine transformations $\mathbf{X}$ via the following optimization:
\begin{equation}
\min_{\mathbf{X}}~ E_{\text{align}}(\mathbf{X}) + \alpha E_{\text{reg}}(\mathbf{X}) + \beta E_{\text{rot}}(\mathbf{X}).
\label{eq:model}
\end{equation}
Here the terms $E_{\text{align}}$, $E_{\text{smooth}}$, and $E_{\text{rot}}$ measures the alignment error, the regularization of transformations across the surface, and the deviation between the transformation matrices and rotation matrices, respectively. $\alpha$ and $\beta$ are positive weights that control the tradeoff between these terms. The definition for each term is explained in the following.

\paragraph{Alignment Term.}
For each transformed source point $\newpos{\mathbf{v}}_i$, we can find the closest target point $\mathbf{u}_{\rho(i)} \in \mathcal{U}$. The alignment term should penalize the deviation between $\newpos{\mathbf{v}}_i$ and $\mathbf{u}_{\rho(i)}$.
A simple approach is to define it as the sum of squared distances between all such pairs of points. This is indeed the alignment error metric used in the classical ICP algorithm~\cite{BeslM92} for rigid registration~\cite{Mitra2004}. On the other hand, such $\ell_2$-norm of pointwise distance can lead to erroneous alignment on real-world data, where the two surfaces might only overlap partially and their point positions might be noisy. This is because partial overlaps and noisy data can induce large distance from some source points to their corresponding target points under the ground-truth alignment, which would be prohibited by the $\ell_2$-norm minimization.

Some previous work, such as the Sparse ICP algorithm from~\cite{Bouaziz2013-SparseICP}, adopts the $\ell_p$-norm $(0 < p < 1)$ as the error metric. It is less sensitive to noises and partial overlaps, since $\ell_p$-norm minimization allows for large distances at some points. However, numerical minimization of the $\ell_p$-norm can be much more expensive than the $\ell_2$-norm. For example, the problem is solved in~\cite{Bouaziz2013-SparseICP} with an iterative algorithm that alternately updates the point correspondence and the transformation which is similar with classical ICP. However, its transformation update has to be done with an inner ADMM solver which is much slower than the closed-form update in classical ICP, and also lack convergence guarantee due to the non-convexity of the problem.

\begin{figure}[!t]
	\centering
	\includegraphics[width=1\columnwidth]{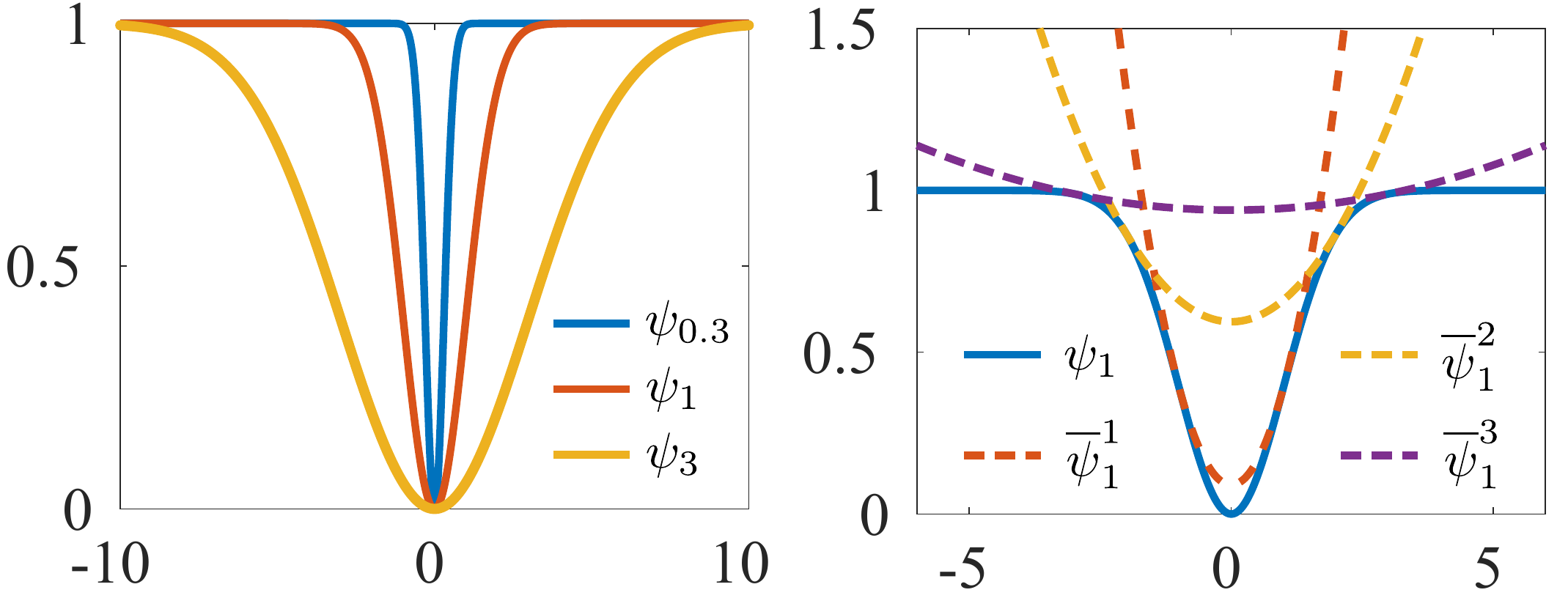}
	\caption{Left: the Welsch's function with different $\nu$ values. Right: different surrogate functions for the Welsch's function with $\nu=1$.}
	\label{fig:welsch}
\end{figure}

Inspired by the recent work from~\cite{ham2015robust} on robust image filtering and~\cite{zhang2018static} on robust mesh filtering, we adopt the following robust metric for the alignment error:
\begin{equation}
E_{\text{align}}(\mathbf{X}) = \sum_{i=1}^{n} \psi_{\nua}(\|\newpos{\mathbf{v}}_i -\mathbf{u}_{\rho(i)}\|),
\label{eq:Ealign}
\end{equation}
where $\psi_{\nua}(\cdot)$ is the Welsch's function~\cite{holland1977robust} (see Fig.~\ref{fig:welsch} left):
\[
    \psi_{\nua}(x)=1-\exp \left(-\frac{x^{2}}{2 \nua^{2}}\right),
\]
and $\nua > 0$ is a user-specified parameter.
$\psi_{\nua}$ is monotonically increasing on $[0, +\infty)$, thus $\psi_{\nua}(\|\newpos{\mathbf{v}}_i -\mathbf{u}_{\rho(i)}\|)$ penalizes the deviation between $\newpos{\mathbf{v}}_i$ and $\mathbf{u}_{\rho(i)}$. On the other hand, as $\psi_{\nua} \leq 1$, the deviation only induces a bounded influence on the metric $E_{\text{align}}$. Moreover, if $\nua \to 0$, then $E_{\text{align}}$ approaches the $\ell_0$-norm of the pointwise distance from the source points to their closest points on the target surface. Therefore, this error metric is insensitive to noisy data and partial overlaps.

\paragraph{Regularization Term.}
Ideally, the transformation induced by two neighboring vertices $\mathbf{p}_i, \mathbf{p}_j$ of the deformation graph should be consistent on their overlapping influenced regions. In~\cite{sumner2007embedded}, such consistency is measured at $\mathbf{p}_i$ using the difference of deformation induced by the transformation $\mathbf{X}_i$ at $\mathbf{p}_i$ and the transformation $\mathbf{X}_j$ at  $\mathbf{p}_j$:
\begin{equation}
    \mathbf{D}_{ij} = \mathbf{A}_j (\mathbf{p}_i - \mathbf{p}_j) + \mathbf{p}_j + \mathbf{t}_j - (\mathbf{p}_i + \mathbf{t}_i).
    \label{eq:Dij}
\end{equation}
Ideally, $\mathbf{D}_{ij}$ should be small across the deformation graph. On the other hand, in some cases the optimal deformation may induce large values of $\mathbf{D}_{ij}$ in some regions, such as the joints of a human body. To reduce the magnitudes of $\mathbf{D}_{ij}$ across the graph while allowing for large magnitudes in some regions, we define the regularization term using the Welsch's function on $\|\mathbf{D}_{ij}\|$:
\begin{equation}
    E_{\text{reg}}(\mathbf{X}) = \sum_{i=1}^r
    \sum_{\mathbf{p}_{j} \in \mathcal{N}(\mathbf{p}_{i})} \psi_{\nur}(\|\mathbf{D}_{ij}\|),
    \label{eq:Ereg}
\end{equation}
where $r$ is the number of nodes in $\mathcal{G}$, $\nur > 0$ is a user-specified parameter, and $\mathcal{N}(\mathbf{p}_{i})$ denotes the set of neighboring vertices for $\mathbf{p}_{i}$ in $\mathcal{G}$.

\paragraph{Rotation Matrix Term.} To preserve rigidity on local surface regions during registration, we would like each transformation $\mathbf{X}_i$ to be as close to a rigid transformation as possible. We measure this property using the deviation between the transformation matrix $\mathbf{A}_i$ and its closest projection onto the rotation matrix group $\SOG = \{\mathbf{R} \in \mathbb{R}^{3 \times 3} \mid \mathbf{R}^T \mathbf{R} = \mathbf{I}, \det(\mathbf{R}) > 0\}$, and define $E_{\text{rot}}$ as
\begin{equation}
E_{\text{rot}}(\mathbf{X}) = \sum_{i=1}^r\|\mathbf{A}_i - \proj\nolimits_{\SOG}(\mathbf{A}_i)\|_F^2,
\label{eq:rigid}
\end{equation}
where $\proj{}$ is the projection operator, and $\|\cdot\|_F$ is the Frobenius norm. 

\subsection{Construction of Deformation Graph}
To construct the deformation graph $\mathcal{G}$, we first extract its vertices $\mathcal{V}_{\mathcal{G}}$ by iteratively adding source points as follows. $\mathcal{V}_{\mathcal{G}}$ is initialized as empty. Then we perform PCA on all source points, and sort them based on their projections onto the axis with the largest eigenvalue of the covariance matrix. According to the sorted order, we go through each source point $\mathbf{v}_i$, and add it to $\mathcal{V}_{\mathcal{G}}$ if $\mathcal{V}_{\mathcal{G}}$ is empty or the shortest geodesic distance from $\mathbf{v}_i$ to the points in $\mathcal{V}_{\mathcal{G}}$ is no smaller than the radius parameter $R$. After determining the vertex set $\mathcal{V}_{\mathcal{G}}$, we construct the edge set $\mathcal{E}_{\mathcal{G}}$ by connecting any two vertices whose geodesic distance is smaller than $R$. We compute the geodesic distance using the fast marching method~\cite{kimmel1998computing}. $R$ is set to $5\overline{l}$ by default, where $\overline{l}$ is the average edge length on the source surface. Compared with alternative sampling strategies such as the farthest point sampling method~\cite{moenning2003fast}, our approach results in fewer sample points for the same radius parameter, which reduces the computational cost while achieving similar accuracy. A comparison example is provided in the supplementary material.

\section{Numerical Algorithm}
The target function $E$ for the optimization problem~\eqref{eq:model} is non-linear and non-convex.
Thanks to the use of the Welsch's function, it can be solved efficiently using the MM algorithm~\cite{Lange2004}. Specifically, given the variable values $\mathbf{X}^{(k)}$ in the current iteration, the MM algorithm constructs a surrogate function $\surrog{E}^{\mathbf{X}^{(k)}}$ for the target function $E$ such that
\begin{equation}
\label{eq:SurrogateE}
\begin{aligned}
\surrog{E}^{\mathbf{X}^{(k)}}(\mathbf{X}^{(k)}) & = E(\mathbf{X}^{(k)}),\\
\surrog{E}^{\mathbf{X}^{(k)}}(\mathbf{X}) &\geq E(\mathbf{X}), \quad \forall \mathbf{X} \neq \mathbf{X}^{(k)}.
\end{aligned}
\end{equation}
Then the variables are updated by minimizing the surrogate function
\begin{equation}
    \mathbf{X}^{(k+1)} = \argmin_{\mathbf{X}} \surrog{E}^{\mathbf{X}^{(k)}}(\mathbf{X}).
\end{equation}
In this way, each iteration is guaranteed to decrease the target function, and the iterations are guaranteed to converge to a local minimum regardless of the initialization~\cite{Lange2004}. In comparison, existing solvers for minimizing the non-convex $\ell_p$-norm such as ADMM~\cite{Bouaziz2013-SparseICP} or iteratively reweighted least squares~\cite{Daubechies2010} either lack convergence guarantee or rely on strong assumptions for convergence.  In the following, we explain the construction of the surrogate function, and present a numerical algorithm for its minimization in each iteration.

\subsection{Surrogate Function}
\label{sec:surrogate}
To construct the surrogate function $\surrog{E}^{\mathbf{X}^{(k)}}$, we note that there is a convex quadratic surrogate function for the Welsch's function $\psi_{\nu}$ at $y$~\cite{ham2015robust} (see Fig.~\ref{fig:welsch} right):
\[
    \surrog{\psi}_{\nu}^y (x) = \psi_{\nu}(y) + \left(\frac{1 - \psi_{\nu}(y)}{ 2 \nu^{2}}\right)\left(x^{2}-y^{2}\right).
\]
This function bounds the Welsch's function from above, and the two function graphs touch at $x = y$.
Applying this surrogate to all relevant terms, we obtain a surrogate function for ${E}_{\textrm{reg}}$ at $\mathbf{X}^{(k)}$. Moreover, we can ignore all constant terms as they do not affect the minimum, resulting in the following convex quadratic function
\begin{equation}
    \surrog{E}_{\textrm{reg}}^{\mathbf{X}^{(k)}}
    = \frac{1}{2 \nur^2} \sum_{i=1}^{r} \sum_{\mathbf{p}_{j} \in \mathcal{N}(\mathbf{p}_{i})} \exp\Big(-\frac{\|\mathbf{D}_{ij}^{(k)}\|^2}{2 \nur^2}\Big) \|\mathbf{D}_{ij}\|^2,
    \label{eq:RegSurrogate}
\end{equation}
where $\mathbf{D}_{ij}^{(k)}$ is evaluated using Eq.~\eqref{eq:Dij} at $\mathbf{X}^{(k)}$. Similarly, we can obtain the following function as a surrogate for ${E}_{\textrm{align}}$ at $\mathbf{X}^{(k)}$ up to a constant:
\begin{equation}
    \frac{1}{2 \nua^2} \sum_{i=1}^{n} \exp\Big(-\frac{\|\newpos{\mathbf{v}}_i^{(k)} -\mathbf{u}_{\rho(i)}^{(k)}\|^2}{2 \nua^2}\Big) \|\newpos{\mathbf{v}}_i -\mathbf{u}_{\rho(i)}\|^2,
    \label{eq:NaiveSurrogate}
\end{equation}
where $\newpos{\mathbf{v}}_i^{(k)}$ is the transformed position of $\mathbf{v}_i$ according to $\mathbf{X}^{(k)}$, and $\mathbf{u}_{\rho(i)}^{(k)}$ is the closest target point for $\newpos{\mathbf{v}}_i^{(k)}$. Eq.~\eqref{eq:NaiveSurrogate} is not a quadratic function of $\mathbf{X}$, as $\mathbf{u}_{\rho(i)}$ depends non-linearly on $\mathbf{X}$. To obtain a more simple form, we note that the term $\|\newpos{\mathbf{v}}_i -\mathbf{u}_{\rho(i)}\|^2$ has a quadratic surrogate function $\|\newpos{\mathbf{v}}_i -\mathbf{u}_{\rho(i)}^{(k)}\|^2$ at $\mathbf{X}^{(k)}$. Applying it to Eq.~\eqref{eq:NaiveSurrogate}, we obtain the following convex quadratic surrogate function for ${E}_{\textrm{align}}$  at $\mathbf{X}^{(k)}$ up to a constant:
\begin{equation}
\surrog{E}_{\textrm{align}}^{\mathbf{X}^{(k)}} = \frac{1}{2 \nua^2} \sum_{i=1}^{n} \exp\Big(-\frac{\|\newpos{\mathbf{v}}_i^{(k)} -\mathbf{u}_{\rho(i)}^{(k)}\|^2}{2 \nua^2}\Big) \|\newpos{\mathbf{v}}_i -\mathbf{u}_{\rho(i)}^{(k)}\|^2.
\label{eq:AlignSurrogate}
\end{equation}
Replacing ${E}_{\textrm{align}}$ and ${E}_{\textrm{reg}}$ in the target function with their surrogates, we arrive at the following MM iteration scheme:
\begin{equation}
    \mathbf{X}^{(k+1)} = \min_{\mathbf{X}} \surrog{E}^{\mathbf{X}^{(k)}} (\mathbf{X}),
    \label{eq:MMIeration}
\end{equation}
where
$\surrog{E}^{\mathbf{X}^{(k)}} = \surrog{E}_{\text{align}}^{\mathbf{X}^{(k)}} + \alpha \surrog{E}_{\text{reg}}^{\mathbf{X}^{(k)}} + \beta E_{\text{rot}}$.

\subsection{Numerical Minimization}

The target function $\surrog{E}^{\mathbf{X}^{(k)}}$ in Eq.~\eqref{eq:MMIeration} still contains a non-linear term $E_{\text{rot}}$, as the projection onto the rotation matrix group depends non-linearly on $\mathbf{A}_i$. On the other hand, as explained later, the special structure of $E_{\text{rot}}$ leads to a  simple form of its gradient, allowing us to evaluate the gradient of $\surrog{E}^{\mathbf{X}^{(k)}}$ efficiently. Therefore, we solve the sub-problem~\eqref{eq:MMIeration} using an L-BFGS solver for fast convergence. In each iteration, L-BFGS utilizes the gradients of $\surrog{E}^{\mathbf{X}^{(k)}}$ at the latest $m+1$ iterates $\mathbf{X}_{(j)}, \mathbf{X}_{(j-1)}, \ldots, \mathbf{X}_{(j-m)}$ to implicitly approximate its inverse Hessian and derive a descent direction $\mathbf{d}_{(j)}$, followed by a line search along $\mathbf{d}_{(j)}$ for a new iterate $\mathbf{X}_{(j+1)}$ with sufficient decrease of the target function~\cite{Nocedal2006}.
In the following, we present the details of the solver.

\paragraph{Gradient Computation.}
For the function $E_{\text{rot}}$ defined in Eq.~\eqref{eq:rigid}, each term $\|\mathbf{A}_i - \proj\nolimits_{\SOG}(\mathbf{A}_i)\|_F^2$ is the squared Euclidean distance from the matrix $\mathbf{A}_i$ to the manifold ${\SOG}$ of rotation matrices. Even though $\proj\nolimits_{\SOG}(\mathbf{A}_i)$ depends non-linearly on $\mathbf{A}_i$,  it can be shown that the squared distance has a simple form of gradient~\cite{Gomes2003}:
\[
    \frac{\partial \| \mathbf{A}_i - \proj\nolimits_{\SOG}(\mathbf{A}_i) \|^2}{\partial \mathbf{A}_i} = 2 \left(\mathbf{A}_i - \proj\nolimits_{\SOG}(\mathbf{A}_i)\right).
\]
Thus we can write the gradient of  $E_{\text{rot}}$ as:
\[
    \frac{\partial E_{\text{rot}}}{\partial \mathbf{A}_i} =  2 \left(\mathbf{A}_i - \proj\nolimits_{\SOG}(\mathbf{A}_i)\right),
    \quad
    \frac{\partial E_{\text{rot}}}{\partial \mathbf{t}_i} = \mathbf{0}.
\]
Since $\surrog{E}_{\text{align}}^{\mathbf{X}^{(k)}}$ and $\surrog{E}_{\text{reg}}^{\mathbf{X}^{(k)}}$ are quadratic functions, their gradients have simple linear forms. To facilitate presentation, we first rewrite the functions in matrix forms. In the following, we assume all 3D vectors are $3 \times 1$ matrices. The transformation at $\mathbf{p}_i$ is represented as $\mathbf{X}_i = [\mathbf{A}_i, \mathbf{t}_i]^T \in \mathbb{R}^{4 \times 3}$, and $\mathbf{X} = [ \mathbf{X}_1^T, \ldots, \mathbf{X}_r^T  ]^T \in \mathbb{R}^{4r \times 3}$. Then we have
\begin{equation}
     \surrog{E}_{\text{align}}^{\mathbf{X}^{(k)}}=\|\mathbf{W}_a (\mathbf{F}\mathbf{X}+\mathbf{P}-\mathbf{U})\|_F^2,
    \label{eq:alignMatrixForm}
\end{equation}
where $\mathbf{W}_a = \text{diag}(\sqrt{w_1^a},...,\sqrt{w_n^a}) \in \mathbb{R}^{n \times n}$ with
$
w_i^a = \frac{1}{2 \nua^2}  \exp\Big(-\frac{\|\newpos{\mathbf{v}}_i^{(k)} -\mathbf{u}_{\rho(i)}^{(k)}\|^2}{2 \nua^2}\Big)
$,
$\mathbf{F}$ is a block matrix  $\{\mathbf{F}_{ij}\}_{1 \leq i \leq n \atop 1 \leq j \leq r} \in \mathbb{R}^{n \times 4r}$ with
\[
\mathbf{F}_{ij}
    = \left\{
    \begin{array}{ll}
    w_{ij} \cdot [ \mathbf{v}_i^T - \mathbf{p}_j^T, 1 ] & \textrm{if}~\mathbf{p}_j \in \mathcal{I}(\mathbf{v}_i)\\
    \mathbf{0} & \textrm{otherwise}
    \end{array}
    \right.,
\]
and
$
    \mathbf{P}= \left[\sum_{\mathbf{p}_j \in \mathcal{I}(\mathbf{v}_i)} \mathbf{p}_j, \ldots, \sum_{\mathbf{p}_j \in \mathcal{I}(\mathbf{v}_n)} \mathbf{p}_j\right]^T \in \mathbb{R}^{n \times 3}
$,
$\mathbf{U}  = \left[ \mathbf{u}_{\rho(1)}^{(k)},..., \mathbf{u}_{\rho(n)}^{(k)}\right]^T \in \mathbb{R}^{n \times 3}.
$
Similarly, we have
\begin{equation}
\surrog{E}_{\text{reg}}^{\mathbf{X}^{(k)}} = \|\mathbf{W}_r(\mathbf{B} \mathbf{X} -\mathbf{Y})\|_F^2,
\label{eq:regMatrixForm}
\end{equation}
where the matrices $\mathbf{B} \in \mathbb{R}^{2 |\mathcal{E}_{\mathcal{G}}| \times 4 r}$ and $\mathbf{Y} \in \mathbb{R}^{2 |\mathcal{E}_{\mathcal{G}}| \times 3}$ encode the computation of a term $\mathbf{D}_{ij}$ in each row, and the diagonal matrix $\mathbf{W}_r$ stores the weight $\sqrt{\frac{1}{2 \nur^2}\exp\Big(-\frac{\|\mathbf{D}_{ij}^{(k)}\|^2}{2 \nur^2}\Big)}$ at the corresponding row.
In the row corresponding to $\mathbf{D}_{ij}$, the elements in $\mathbf{Y}$ are $[\mathbf{p}_j^T - \mathbf{p}_i^T]$, whereas the elements in $\mathbf{B}$ for $\mathbf{X}_i$ and $\mathbf{X}_j$ are $[\mathbf{p}_i^T - \mathbf{p}_j^T, 1]$ and $[0, 0, 0, 1]$, respectively.
We can further write the gradient of $E_{\text{rot}}$ in matrix form as
\[
    \frac{\partial E_{\text{rot}}}{\partial \mathbf{X}}
    = 2 (\mathbf{J} \mathbf{X} - \mathbf{Z}),
\]
where $\mathbf{Z} = [ \proj_{\mathcal{R}}(\mathbf{A}_1), \mathbf{0}, \ldots, \proj_{\mathcal{R}}(\mathbf{A}_r), \mathbf{0} ]^T \in \mathbb{R}^{4r \times 3}$, and $\mathbf{J} = \textrm{diag} (1, 1, 1, 0, 1, 1, 1, 0, \ldots, 1, 1, 1, 0) \in \mathbb{R}^{4r \times 4r}$.
We can then derive the gradient function of $\surrog{E}^{\mathbf{X}^{(k)}}$ as
\begin{equation}
\begin{aligned}
    \mathbf{G}(\mathbf{X})
     =~&2 [\mathbf{F}^T \mathbf{W}_a^2 (\mathbf{F}\mathbf{X}+\mathbf{P}-\mathbf{U}) \\
     &+
     \alpha \mathbf{B}^T \mathbf{W}_r^2(\mathbf{B} \mathbf{X} -\mathbf{Y})
        +
     \beta (\mathbf{J} \mathbf{X} - \mathbf{Z})].
\end{aligned}
\label{eq:Gradient}
\end{equation}

\begin{algorithm}[t!]
	%\LinesNumbered
	\label{Alg:TwoLoopRecursion}
	\caption{Two-loop recursion for computing descent direction $\mathbf{d}_{(j)}$}
	$\mathbf{Q} = - \mathbf{G}(\mathbf{X}_{(j)})$\;
	\For{$i = j-1, \ldots, j- m$}{
		$\mathbf{S}_i = \mathbf{X}_{i+1} - \mathbf{X}_i$;~
		$\mathbf{T}_i = \mathbf{G}(\mathbf{X}_{i+1}) - \mathbf{G}(\mathbf{X}_i)$\;
		$\rho_i = \Tr(\mathbf{T}_i^T \mathbf{S}_i)$\;
		$\xi_i = \Tr(\mathbf{S}_i^T \mathbf{Q}) / \rho_i$\;
		$\mathbf{Q} = \mathbf{Q} - \xi_i\mathbf{T}_i $\;
		
	}
	$\mathbf{R} = \mathbf{H}_0^{-1} \mathbf{Q}$\;
	\For{$i = j-m, \ldots, j- 1$}{
		$\eta = \Tr(\mathbf{T}_i^T \mathbf{R}) / \rho_i $\;
		$\mathbf{R} = \mathbf{R} + \mathbf{S}_i(\xi_i - \eta)$\;
	}
	$\mathbf{d}_{(j)} = \mathbf{R}$\;
\end{algorithm}

\paragraph{Computing $\mathbf{d}_{(j)}$ and $\mathbf{X}_{(j+1)}$.}
We adopt the L-BFGS implementation from~\cite{liu2017quasi} to utilize the special structure of $\surrog{E}^{\mathbf{X}^{(k)}}$.
Specifically, given an initial approximation $\mathbf{H}_{0}$ for the Hessian of $\surrog{E}^{\mathbf{X}^{(k)}}$ at $\mathbf{X}_{(j)}$,
we compute the descent direction $\mathbf{d}_{(j)}$ using a two-loop recursion explained in~\cite{liu2017quasi} (see Algorithm~\ref{Alg:TwoLoopRecursion}).
Following~\cite{liu2017quasi}, we derive $\mathbf{H}_{0}$ by assuming a fixed projection $\proj_{\mathcal{R}}(\mathbf{A}_i)$ for the term $E_{\text{rot}}$, resulting in the following approximation:
\begin{equation}
    \mathbf{H}_0 = 2 (\mathbf{F}^T \mathbf{W}_a^2 \mathbf{F}
    + \alpha \mathbf{B}^T \mathbf{W}_r^2\mathbf{B}
    +  \beta \mathbf{J}).
    \label{eq:InitialHessian}
\end{equation}
The new iterate $\mathbf{X}_{(j+1)}$ is then computed as
\[
    \mathbf{X}_{(j+1)} = \mathbf{X}_{(j)} + \lambda \mathbf{d}_{(j)},
\]
using a line search to determine the step size $\lambda > 0$ that achieve sufficient decrease of $\surrog{E}^{\mathbf{X}^{(k)}}$:
\begin{equation}
    \surrog{E}^{\mathbf{X}^{(k)}}(\mathbf{X}_{(j+1)})
    \leq
    \surrog{E}^{\mathbf{X}^{(k)}}(\mathbf{X}_{(j)}) +
    \gamma \lambda \Tr(\left(\mathbf{G}(\mathbf{X}_{(j)})\right)^T \mathbf{d}_{(j)}),
    \label{eq:SufficientDecrease}
\end{equation}
with $\gamma  \in (0, 1)$.
The iterative L-BFGS solver is terminated if $\surrog{E}^{\mathbf{X}^{(k)}}(\mathbf{X}_{(j)}) - \surrog{E}^{\mathbf{X}^{(k)}}(\mathbf{X}_{(j+1)}) < \epsilon_1$ where $\epsilon_1$ is a user-specified threshold. And the outer MM iteration is terminated if $\max_i\|\newpos{\mathbf{v}}^{(k+1)}_i - \newpos{\mathbf{v}}^{(k)}_i\| < \epsilon_2$ with a user-specified threshold $\epsilon_2$ or the number of outer iterations reaches a threshold $I_{\max}$.
In all experiments, we set $m = 5$, $\gamma = 0.3$, $\epsilon_1 = 10^{-3}$, $\epsilon_2 = 10^{-3}$, and $I_{\max} = 100$.

The matrix $\mathbf{H}_0$ is sparse symmetric positive definite and remains unchanged in each iteration of an L-BFGS solver. To solve the linear equation $\mathbf{R} = \mathbf{H}_0^{-1} \mathbf{Q}$ efficiently in the two-loop recursion, we compute a sparse Cholesky factorization for $\mathbf{H}_0$ at the beginning of an L-BFGS run, and reuse it in each iteration to solve the linear system with different right-hand-sides. In addition, the columns of the right-hand-side matrix $\mathbf{Q}$ are independent, thus we solve them in parallel. Moreover, although the matrix $\mathbf{H}_0$ may change between different L-BFGS runs, its sparsity pattern remains the same. Therefore, we pre-compute a symbolic factorization of $\mathbf{H}_0$ and only perform numerical factorization subsequently.

\paragraph{Choosing $\nua$ and $\nur$.}
To achieve robust registration, the values of $\nua$ and $\nur$ play an important role. Each of them acts as the standard deviation parameter for the Gaussian weight in the surrogate function~\eqref{eq:RegSurrogate} or \eqref{eq:AlignSurrogate}. The Gaussian weight becomes effectively zero for error terms whose current magnitude is much larger than the standard deviation. Both $\nua$ and $\nur$ need to be sufficiently small at the final stage of the solver, to exclude the influence of large error terms that arise from partial overlaps etc. On the other hand, at the initial stage of the solver, their values need to be larger in order to accommodate more error terms and achieve coarse alignment. Therefore, we initialize the variables $\mathbf{X}$ with a rigid transformation (see Sec.~\ref{sec:Results} for details), and solve the problem~\eqref{eq:model} with relatively large values $\nua = \nua^{\max}$ and $\nur = \nur^{\max}$. The solution is then used as initial variable values to re-solve the problem~\eqref{eq:model} with the values of $\nua$ and $\nur$ decreased by half. We repeat this process until the value of $\nua$ reaches a lower bound $\nua^{\min}$, and take the final solution as the registration result. Algorithm~\ref{Alg:total-alg} summarizes our overall registration algorithm with gradually decreased values of $\nua$ and $\nur$.
By default, we set $\nur^{\max} = 40 \overline{l}$, $\nua^{\max} = 10 \overline{d}$ and $\nua^{\min} = 0.5 \overline{l}$, where $\overline{d}$ is the medium distance from source points to their corresponding target points under the initial rigid transformation, and $\overline{l}$ is the average edge length on the source surface. In the supplementary material, we compare our strategy of gradually decreasing $\nua$ and $\nur$ with registration using fixed $\nua$ and $\nur$, which shows that our strategy achieves a more accurate result.

\begin{algorithm}[!t]
\LinesNumbered
\label{Alg:total-alg}
\caption{Non-rigid registration}
$\nu_a=\nu_a^{\max};~~\nu_r = \nu_r^{\max}$\;
%;~~$k=0$
\While{TRUE}{
	%$k_{\max} = k + I_{\max}$\;
    $k=0$\;
    \Repeat{$\max_i\|\newpos{\mathbf{v}}^{(k+1)}_i - \newpos{\mathbf{v}}^{(k)}_i\| < \epsilon_2$ OR $k = I_{\max}$}{
        Find correspoding point $\mathbf{u}_{\rho(i)}^{(k)}$ for each $\newpos{\mathbf{v}}_i^{(k)}$\;
        Update weight matrices $\mathbf{W}_{a}$ and $\mathbf{W}_r$\;
        $j = -1$;~~ $\mathbf{X}_{(0)} = \mathbf{X}^{(k)}$\;
         \Repeat{$\surrog{E}^{\mathbf{X}^{(k)}}(\mathbf{X}_{(j)}) - \surrog{E}^{\mathbf{X}^{(k)}}(\mathbf{X}_{(j+1)}) < \epsilon_1$}{
        $j = j+1$\;
        Compute gradient $\mathbf{G}(\mathbf{X}_{(j)})$ with~\eqref{eq:Gradient}\;
        Compute direction $\mathbf{d}_{(j)}$ with Alg.~\ref{Alg:TwoLoopRecursion}\;
        Perform line search for $\mathbf{X}_{(j+1)}$ that satisfies condition~\eqref{eq:SufficientDecrease}\;
        }
        $\mathbf{X}^{(k+1)}=\mathbf{X}_{(j+1)}$\;
        $k = k+1$\;
    }
    \If{$\nu_a = \nu_a^{\min}$}
    {
        return $\mathbf{X}^{(k+1)}$\;
    }
    $\nu_a =\max(0.5 \cdot \nu_a, \nu_a^{\min})$;~~
    $\nu_r = 0.5 \cdot \nu_r$\;
}
\end{algorithm}

\paragraph{Setting of $\alpha$ and $\beta$.} To account for the number of terms in each component of the target function in Eq.~\eqref{eq:model}, we set the weights as $\alpha=k_\alpha (|\mathcal{V}|/|\mathcal{E}_{\mathcal{G}}|)$ and $\beta= k_\beta (|\mathcal{V}|/|\mathcal{V}_{\mathcal{G}}|) $, with  $k_\alpha = k_\beta = 1$ by default. We decrease the values of $k_\alpha$ and $k_\beta$ on problem instances with reliable initialization to achieve better alignment, and increase their values on problem instances with less reliable initialization to avoid converging to an undesirable local minimum. Moreover, the value of $k_\alpha$ can be increased for smoother deformation of the source surface, or decreased for faster convergence.

\section{Results}
\label{sec:Results}
In this section, we evaluate the effectiveness of our approach, and compare its performance with state-of-the-art methods including N-ICP~\cite{amberg2007optimal}, RPTS~\cite{li2018robust}, and SVR-$\ell_0$~\cite{guo2015robust}. The evaluations are performed on the MPI Faust dataset~\cite{Bogo:CVPR:2014} and the human motion datasets from~\cite{vlasic2008articulated}. We only show representative results in this section, with more comparisons provided in the supplementary material.

\begin{figure}[!t]
	\centering
	 \includegraphics[width=\columnwidth]{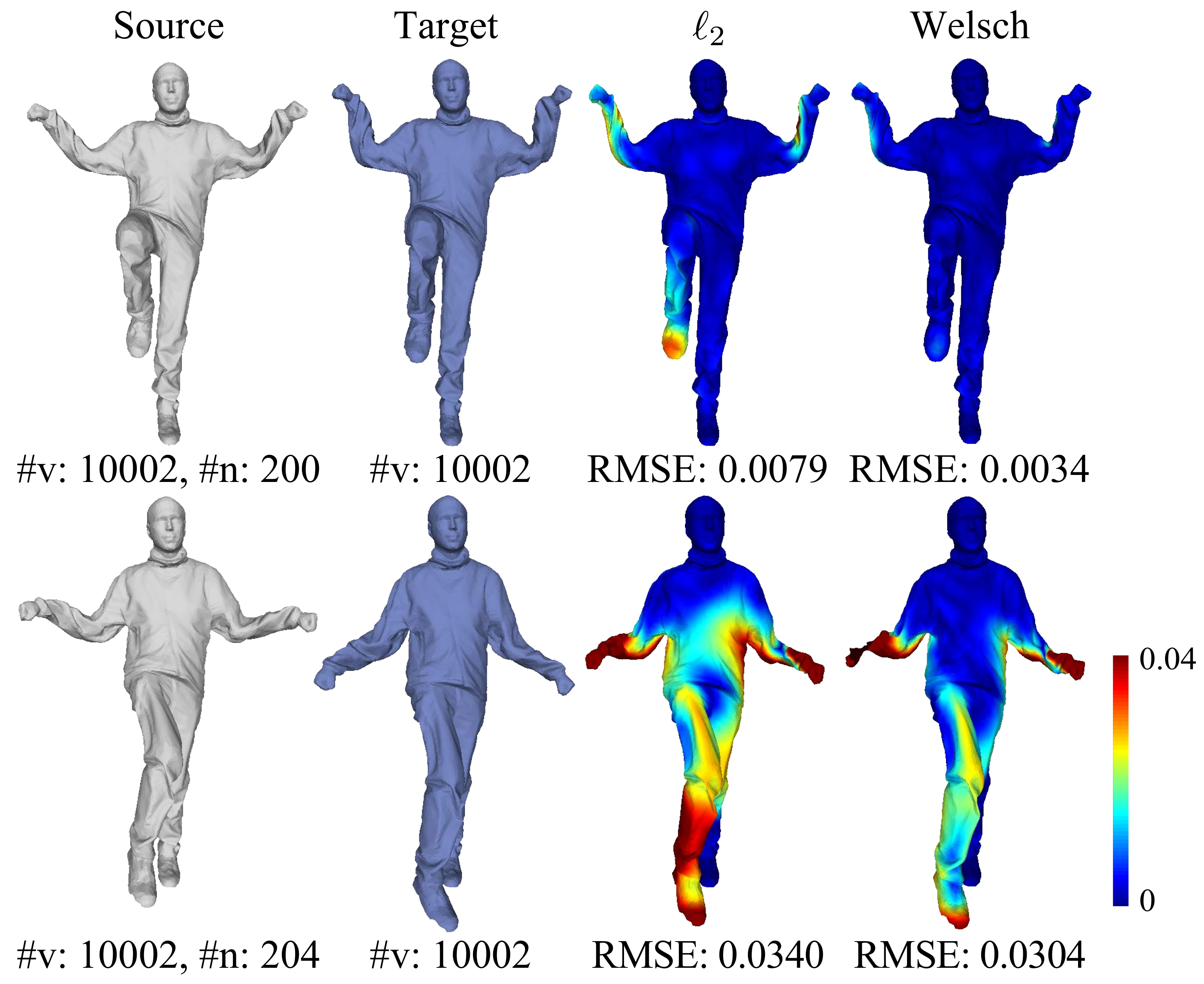}
	\caption{Comparison between our formulation and an alternative formulation using the $\ell_2$-norm instead of the Welsch's function, tested on two pairs of meshes from the ``crane'' dataset of~\cite{vlasic2008articulated}. We set with $k_{\alpha} = 0.1, k_{\beta} = 100$ for the $\ell_2$-norm formulation, and $k_{\alpha} = 1, k_{\beta} = 10^4$ for our formulation.}
\vspace{-5mm}
	\label{fig:compare_l2}
\end{figure}

\paragraph{Implementation Details}
We implement our method in C++, and all results and comparisons are tested on a PC with 16GB of RAM and a 6-core CPU at 3.60GHz.
Each pair of surfaces are pre-processed by aligning their centroids and scaling them to achieve unit-length diagonal for their combined bounding box.
We initialize the transformation variables using a rigid transformation computed via 15 iterations of the ICP algorithm to align the source and target points, with rejection of point pairs whose distance is larger than a threshold $\epsilon_d$ or whose normal vectors deviate by more than an angle $\theta$~\cite{Rusinkiewicz2001}. We set $\epsilon_d = 0.3$ and $\theta = 60^\circ$ by default. The ICP iterations are initialized by aligning corresponding feature points, determined either using the closest point pairs between the pre-processed surfaces coupled with the above rejection criteria, or using the SHOT feature~\cite{tombari2010unique} with diffusion pruning~\cite{tam2014diffusion}, or through manual labeling.
In the figures, we visualize the initial correspondence using blue lines connecting the corresponding points.
We evaluate the registration accuracy via the root mean square error compared with the ground truth:
\begin{equation}
\text{RMSE}=\sqrt{\frac{\sum_{\mathbf{v}_i \in{\mathcal{V}}}e_i^2}{|\mathcal{V}|}},
\end{equation}
where $e_i  = \|\mathbf{v}_i^{\ast} -\mathbf{v}_i^{\text{gt}}\|$
is the deviation between the transformed positions $\mathbf{v}_i^{\ast}, \mathbf{v}_i^{\text{gt}}$ of a source point $\mathbf{v}_i$ under the computed and the ground-truth transformations respectively. In the figures, we show the RMSE for each registration result, and use color-coding to visualize the error values $\{e_i\}$ across the surface. Unless stated otherwise, all RMSE values and color-coded registration errors are in the unit of meters. We use $\#$v to denote the number of sample points on a surface, and $\#$n for the number of deformation graph vertices.
Similar to our formulation~\eqref{eq:model}, the optimization target functions of N-ICP, RPTS, and SVR-$\ell_0$ all involve a regularization term and/or a term for the rigidity of transformations. To ease presentation, for all methods we use $\alpha$ and $\beta$ to denote the weights for the regularization term and the rigidity term, respectively. We tune the weights for each method to obtain their best results for comparison.
In addition, N-ICP and RPTS require dropping unreliable correspondence between source and target points in each iteration, and we adopt the distance and normal deviation thresholds $\epsilon_d$ and $\theta$ as used in the ICP iterations for initialization.

\begin{figure}[!t]
	\centering
	 \includegraphics[width=\columnwidth]{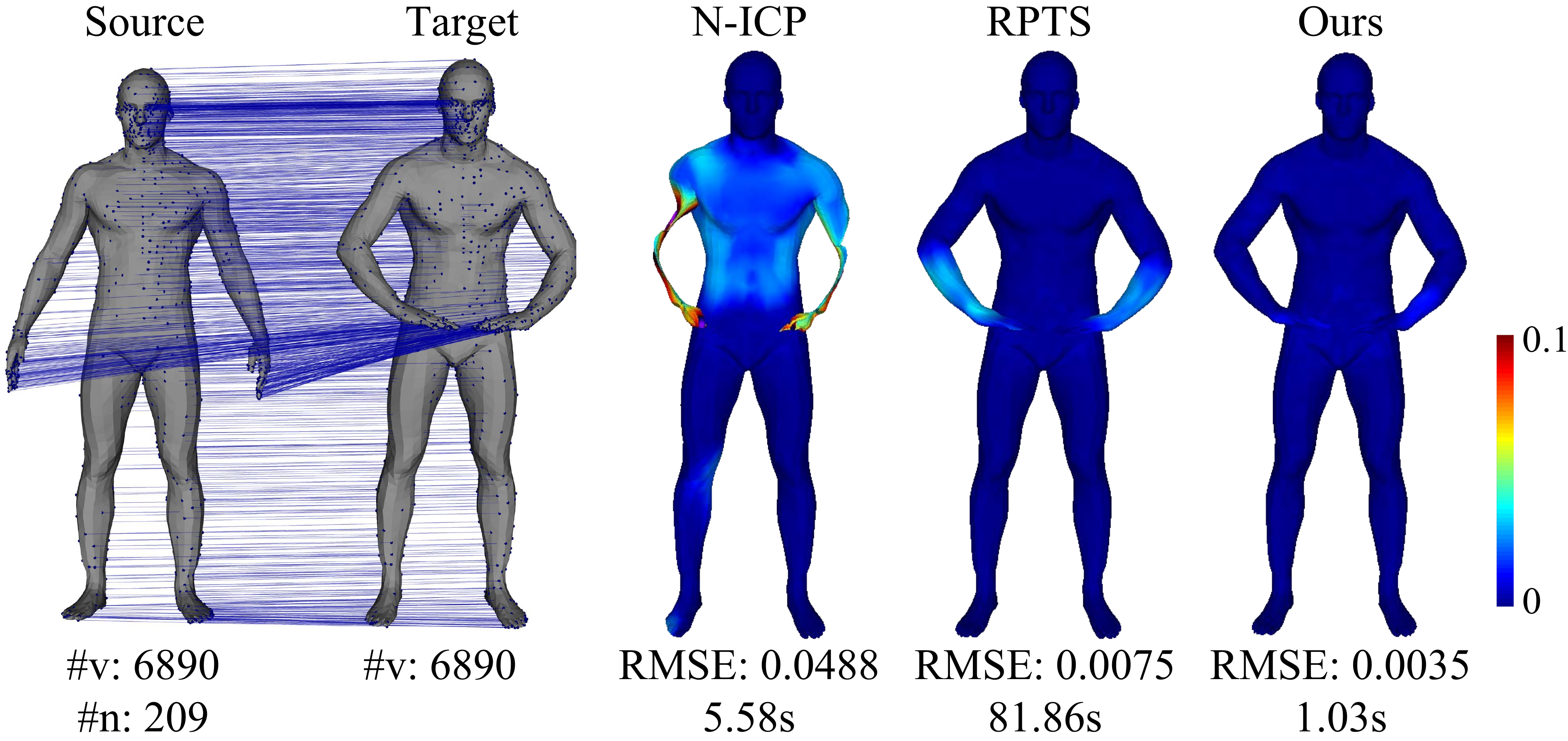}   \caption{Comparison of registration results on an example from the MPI Faust dataset.}
	\label{fig:faust}
\end{figure}

\begin{table}[!t]
	\caption{Average computation time and RMSE (in millimeters) on the MPI Faust dataset. We set $\alpha = 5$ for N-ICP,  $\alpha = 100, \beta = 0.1$ for RPTS, and $k_{\alpha} = k_{\beta} = 0.001$ for our method.}
	\label{Tab:faust}
	\setlength{\tabcolsep}{1.6pt}
	\centering
	\begin{small}
		\begin{tabular}{ | c |  c  c | c  c | c  c | c  c  |}
			\Xhline{1pt}
			\multicolumn{1}{ | c |}{\multirow{2}{*}{Pose Pair}}
			&\multicolumn{2}{ c |}{N-ICP}
			&\multicolumn{2}{ c | }{RPTS}
			&\multicolumn{2}{ c |}{Ours}    \\\cline{2-7}
			%\midrule
			\multicolumn{1}{ | c |}{}
			&Time  (s)     &RMSE
			&Time  (s)     &RMSE
			&Time  (s)     &RMSE \\\hline
	        1 & 7.43  & 59.7  & 61.83  & 9.16  & \textbf{1.24}  & \textbf{5.46} \\
        2 & 4.46  & 20.8  & 46.52  & 3.22  & \textbf{0.86}  & \textbf{1.35} \\
        3 & 7.95  & 80.1  & 35.25  & \textbf{7.97}  & \textbf{1.36}  & 8.99 \\
        4 & 5.45  & 45.4  & 22.99  & 5.69  & \textbf{1.27}  & \textbf{4.26} \\
        5 & 4.13  & 14.2  & 14.04  & \textbf{0.717}  & \textbf{0.81}  & 1.06 \\
        6 & 8.61  & 59.6  & 28.72  & 6.29  & \textbf{1.30}  & \textbf{3.89} \\
        7 & 12.57  & 208  & 43.26  & 61.6  & \textbf{2.23}  & \textbf{49.3} \\
        8 & 6.05  & 70.4  & 26.27  & \textbf{6.58}  & \textbf{1.34}  & 7.54 \\\hline
        Mean & 7.08  & 69.7  & 34.86  & 12.7  & \textbf{1.30}  & \textbf{10.2} \\
        Median & 6.74  & 59.6  & 31.99  & 6.44  & \textbf{1.28}  & \textbf{4.86} \\
			\Xhline{1pt}
		\end{tabular}
	\end{small}
\end{table}

\subsection{Effectiveness of the Welsch's Function}
We perform the optimization~\eqref{eq:model} with the Welsch's functions in $E_{\text{align}}$ and $E_{\text{reg}}$ replaced by square functions. Fig.~\ref{fig:compare_l2} compares the registration accuracy of such $\ell_2$-norm formulation with our approach, on two problem instances from the ``crane'' dataset of~\cite{vlasic2008articulated}. we can see that the Welsch's function leads to more accurate results than the $\ell_2$-norm.

\begin{figure}[!t]
	\centering
	 \includegraphics[width=\columnwidth]{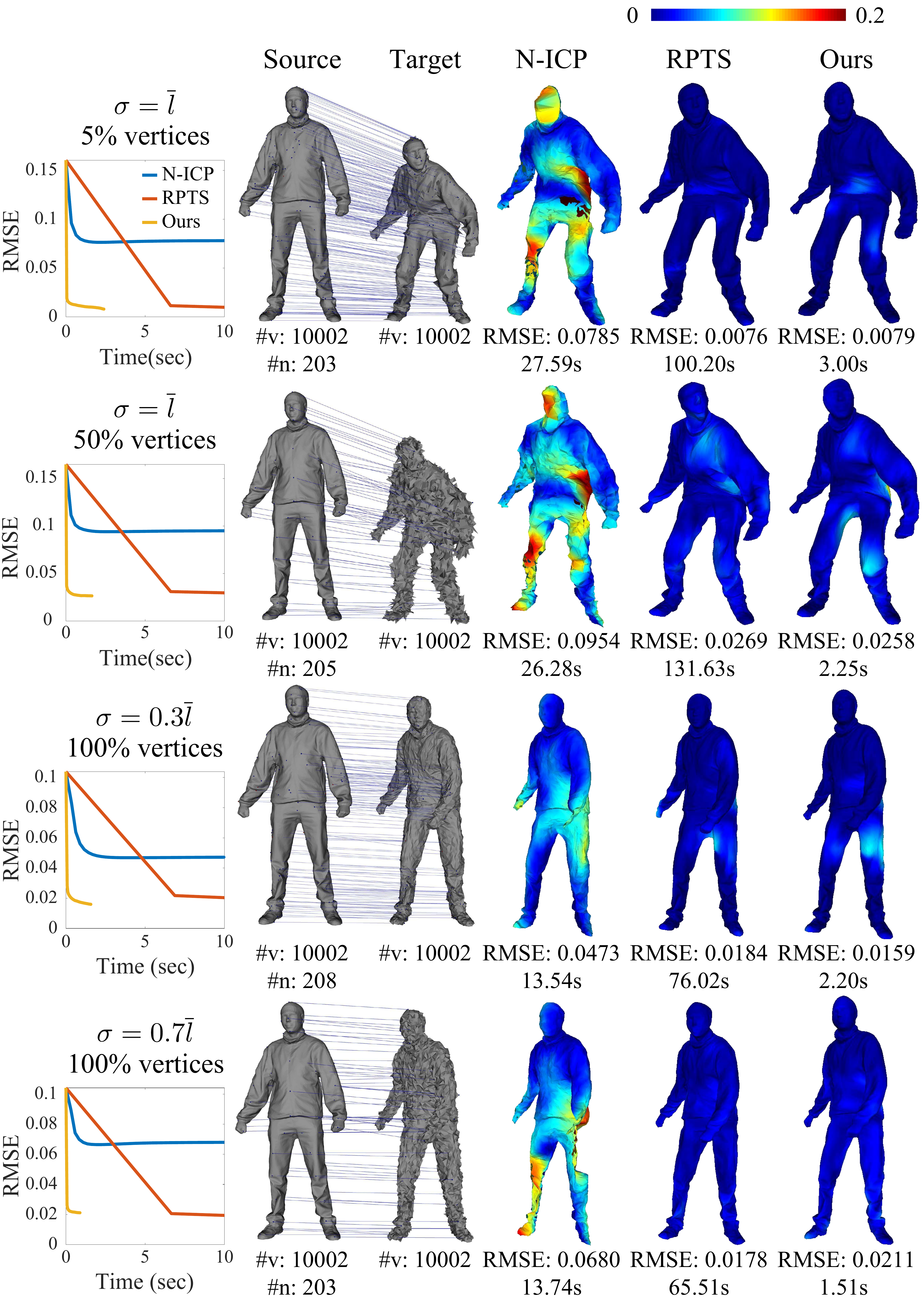}
	\caption{Comparison with N-ICP and RPTS on noisy models constructed by adding Gaussian noise to some vertices on the target surface. The original models are taken from the ``jumping'' dataset of~\cite{vlasic2008articulated}. The left column shows how the RMSE of registration results changes over time during optimization. We set $\alpha = 5$ for N-ICP, $\alpha = 10, \beta = 1$ for RPTS, $k_{\alpha} = 0.01, k_{\beta} = 1$ for our method, and $\theta=180^\circ$ for all methods.}
\vspace{-5mm}
	\label{fig:noise}
\end{figure}

\subsection{Comparison with Other Methods}
In Tab.~\ref{Tab:faust} and Fig.~\ref{fig:faust}, we compare our method with N-ICP and RPTS on the MPI Faust dataset.
We select 10 subjects with 9 poses, and use the first pose of each subject as the source surface the other poses of the subject as the targets.
We evaluate the average RMSE among all subjects for the same pose pair, and list them in Tab.~\ref{Tab:faust}.
Fig.~\ref{fig:faust} shows the results from different methods on a pose pair for a subject. We can see that our method requires significantly less computational time while achieving similar or better accuracy. A major reason for the efficiency of our method is the adoption of a deformation graph, which only requires optimizing one affine transformation per graph vertex. In comparison, N-ICP and RPTS require optimizing one affine transformation per mesh vertex, which significantly increases the number of variables as well as computational time.

Fig.~\ref{fig:noise} compares our method with N-ICP and RPTS on models from the ``jumping'' dataset of~\cite{vlasic2008articulated}, with added Gaussian noise on vertex positions of the target surface along their normal directions.
In the first two rows of Fig.~\ref{fig:noise}, we add noise with standard deviation $\sigma = \overline{l}$ to 5\% or 50\% of the vertices on the target surface respectively, where $\overline{l}$ is the average edge length of the target surface.
In the last two rows, we add noise to all vertices of the target surface with stand deviation $\sigma = 0.3 \overline{l}$ and $\sigma = 0.7 \overline{l}$, respectively.
The comparison shows that our method is more robust to noisy data.

In Fig.~\ref{fig:partial}, we compare our method with  N-ICP, RPTS and SVR-$\ell_0$ on partially overlapping data, which is synthesized from the ``bouncing'' dataset of~\cite{vlasic2008articulated} by removing some vertices from the target surface. For SVR-$\ell_0$, we use the squared distance from source points to their closest target points as the data fitting term. We can see that our method is significantly faster than other methods, and is more robust to such partial overlapping model.

\begin{figure}[!t]
	\centering
	 \includegraphics[width=1\columnwidth]{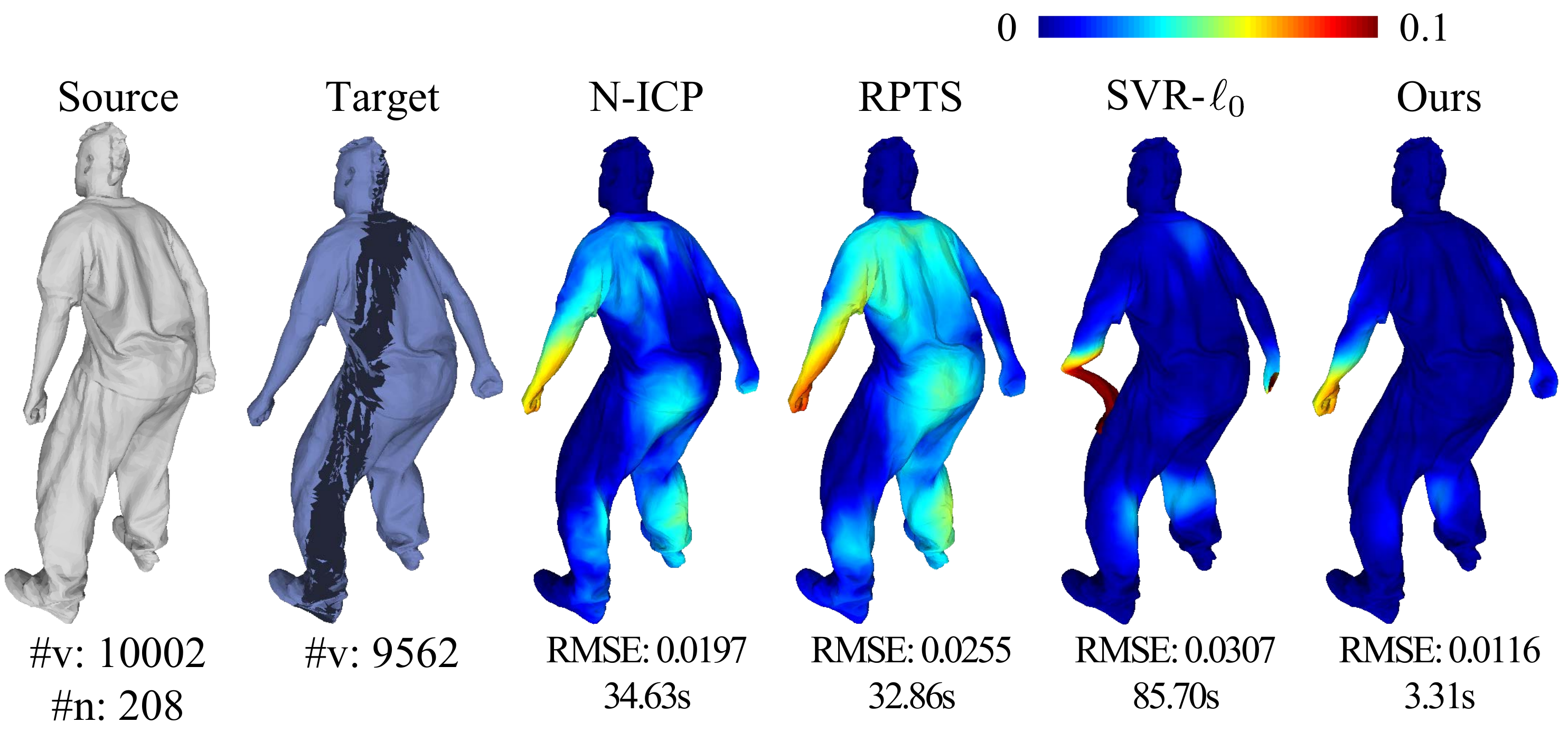}
	\caption{Comparison with N-ICP, RPTS and SVR-$\ell_0$ on models with partial overlaps, constructed by removing some vertices from the target surface. The original models are taken from the ``bouncing'' dataset of~\cite{vlasic2008articulated}. We set $\alpha = 10$ for N-ICPS, $\alpha = 1, \beta=100$ for RPTS, $\alpha = 0.1,\beta = 100$ for SVR-$\ell_0$, $k_{\alpha} = 1, k_{\beta} = 100, \nua^{\max}=30\overline{d}, \nur^{\max} = 100\overline{l}$ for our method, and $ \theta=45^\circ$ for all methods.}
\vspace{-5mm}
	\label{fig:partial}
\end{figure}

\section{Conclusion}

In this paper, we proposed a robust non-rigid registration model based on the Welsch's function. Applying the Welsch's function to the alignment term and the regularization term makes the formulation robust to the noise and partial overlap. To efficiently solve this problem, we apply majorization-minimization to transform the nonlinear and non-convex problem into a sequence of simple sub-problems that are efficiently solved with L-BFGS. Extensive experiments demonstrate the effectiveness of our method and its efficiency compared to existing approaches. 

\noindent \textbf{Acknowledgement}
We thank the authors of~\cite{guo2015robust} for providing their implementation.
This work was supported by the National Natural Science Foundation of China (No. 61672481), and Youth Innovation Promotion Association CAS (No. 2018495).

{\small
\bibliographystyle{ieee_fullname}
\bibliography{egbib}
}
\section*{Appendix}
\subsection*{The choice of sampling radius}
\begin{figure}[!htb]
    \centering
    \includegraphics[width=0.95\columnwidth]{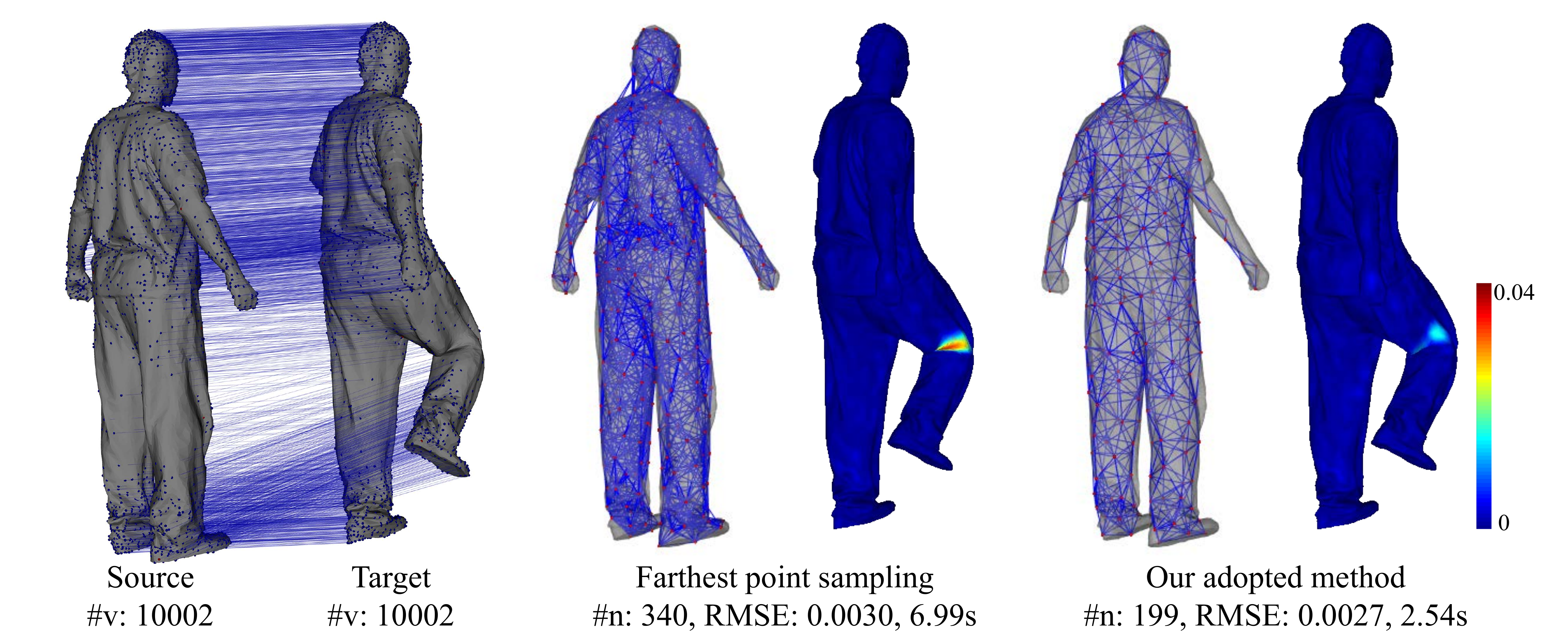}
    \caption{Comparison with the farthest sampling method for the given radius $R=5\overline{l}$, where $\overline{l}$ is the average edge length on the source surface. The farthest point sampling method can obtain more graph nodes.($k_{\alpha}=0.001, k_{\beta}=0.1$). The RMSE and the color-coded registration errors are in the unit of meters.}
    \label{fig:diffsample}
\end{figure}

\begin{figure}[!htb]
    \centering
    \includegraphics[width=\columnwidth]{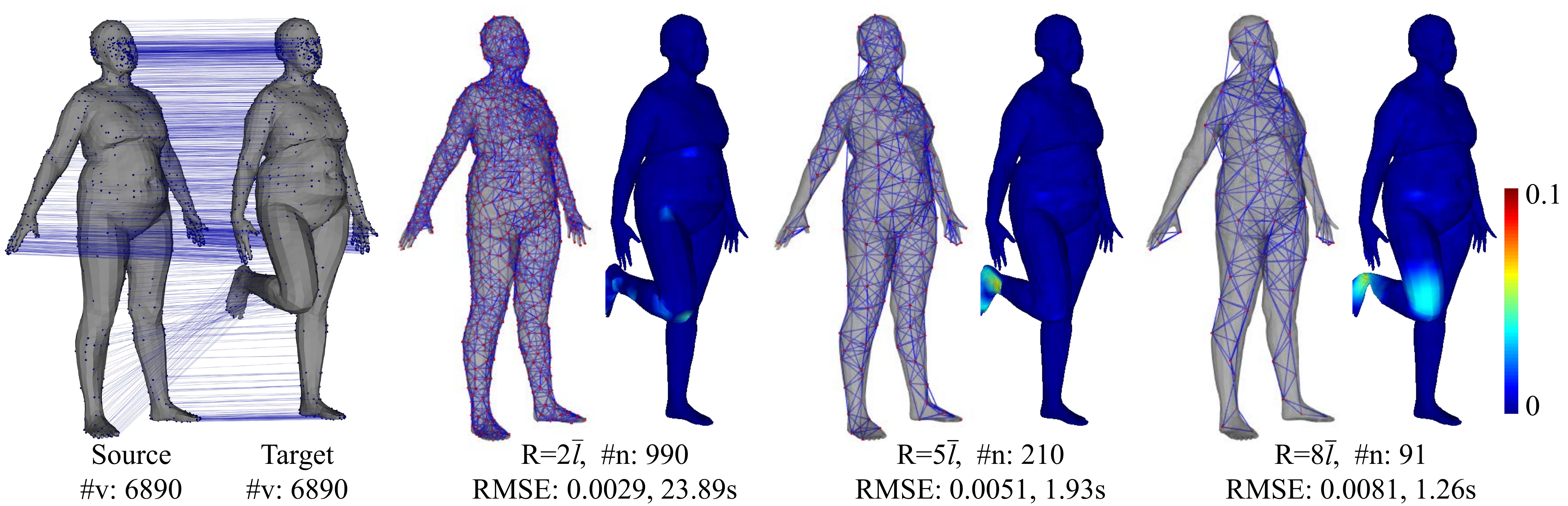}
    \caption{Comparison with different sampling radius with $k_{\alpha}=0.001, k_{\beta}=0.1$. More nodes can get more accurate results. The RMSE and the color-coded registration errors are in the unit of meters.}
    \label{fig:diffr}
\end{figure}

The number of graph nodes and edges will influence the memory footprint and computational cost for the solver. The farthest point sampling method~\cite{moenning2003fast} is to repeatedly add the farthest point to the graph until the geodesic distance between graph nodes and the farthest point is smaller than the given radius parameter $R$. Compared with farthest point sampling method(Fig.~\ref{fig:diffsample}), our adopt method can obtain fewer nodes and is faster to converge with the similar accuracy. In our method, the radius $R$ can be used to balance the speed and accuracy. A smaller $R$ leads to more nodes in the deformation graph, which increases the number of variables and accuracy while requires more computational time. We show the comparison in Fig.~\ref{fig:diffr}. Our method does not vary the sampling density based on curvature. In our experiments, such uniform density is sufficient to generate good results and curvature-adaptive sampling can be a future work.

\subsection*{The comparison with fixed parameters}
\begin{figure}[!htb]
    \centering
    \includegraphics[width=1\columnwidth]{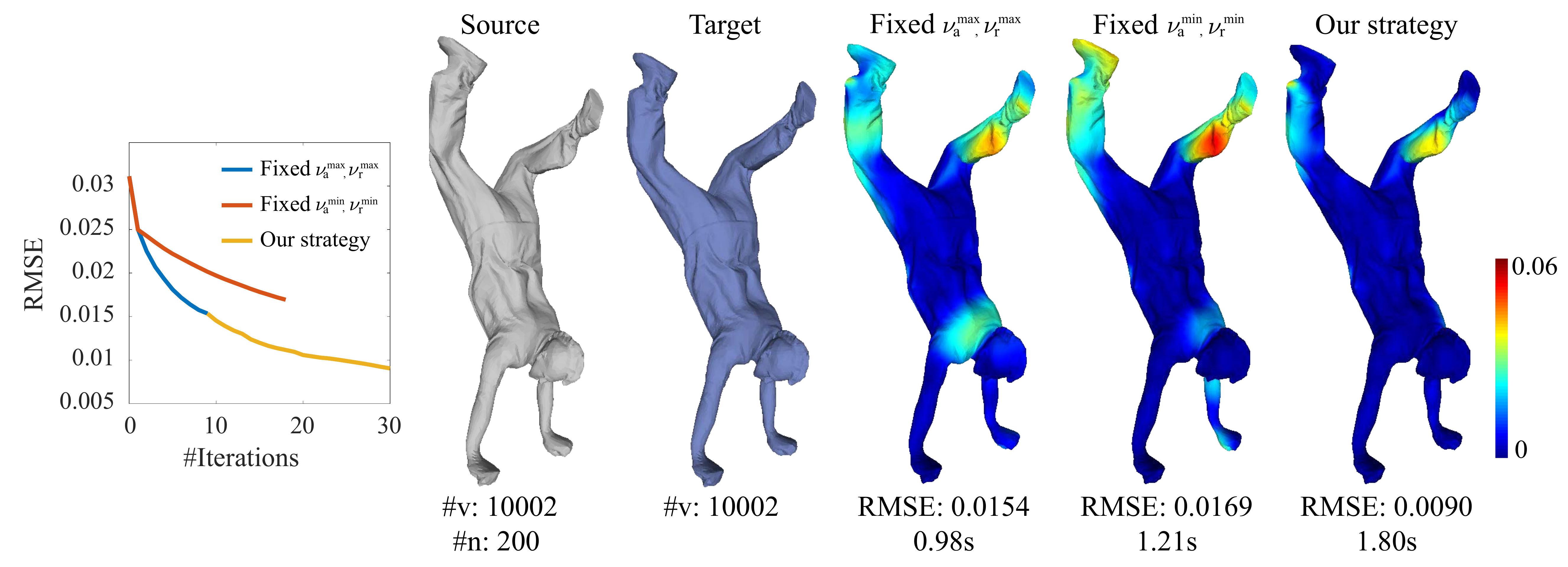}
    \caption{Comparison with fixed $\nua, \nur$ on 42-th to 40-th mesh in "handstand" with $k_{\alpha}=100$ and $k_{\beta}=50$. Here $\nur^{\min}$ is the value when $\nua$ reaches $\nua^{\min}$. The RMSE and the color-coded registration errors are in the unit of meters.}
    \label{fig:fixnu}
\end{figure}

In our method, the $\nua$ and $\nur$ values will influence the registered result, and discussion on this part is given in "Choosing $\nua$ and $\nur$"(Sec 4.2) of the paper. In Fig.~\ref{fig:fixnu}, we show the comparison between our dynamic adjustment strategy and the strategy by fixing $\nua, \nur$, and we can see our method can get higher accuracy.

\begin{figure*}[b]
    \centering
     \includegraphics[width=\textwidth]{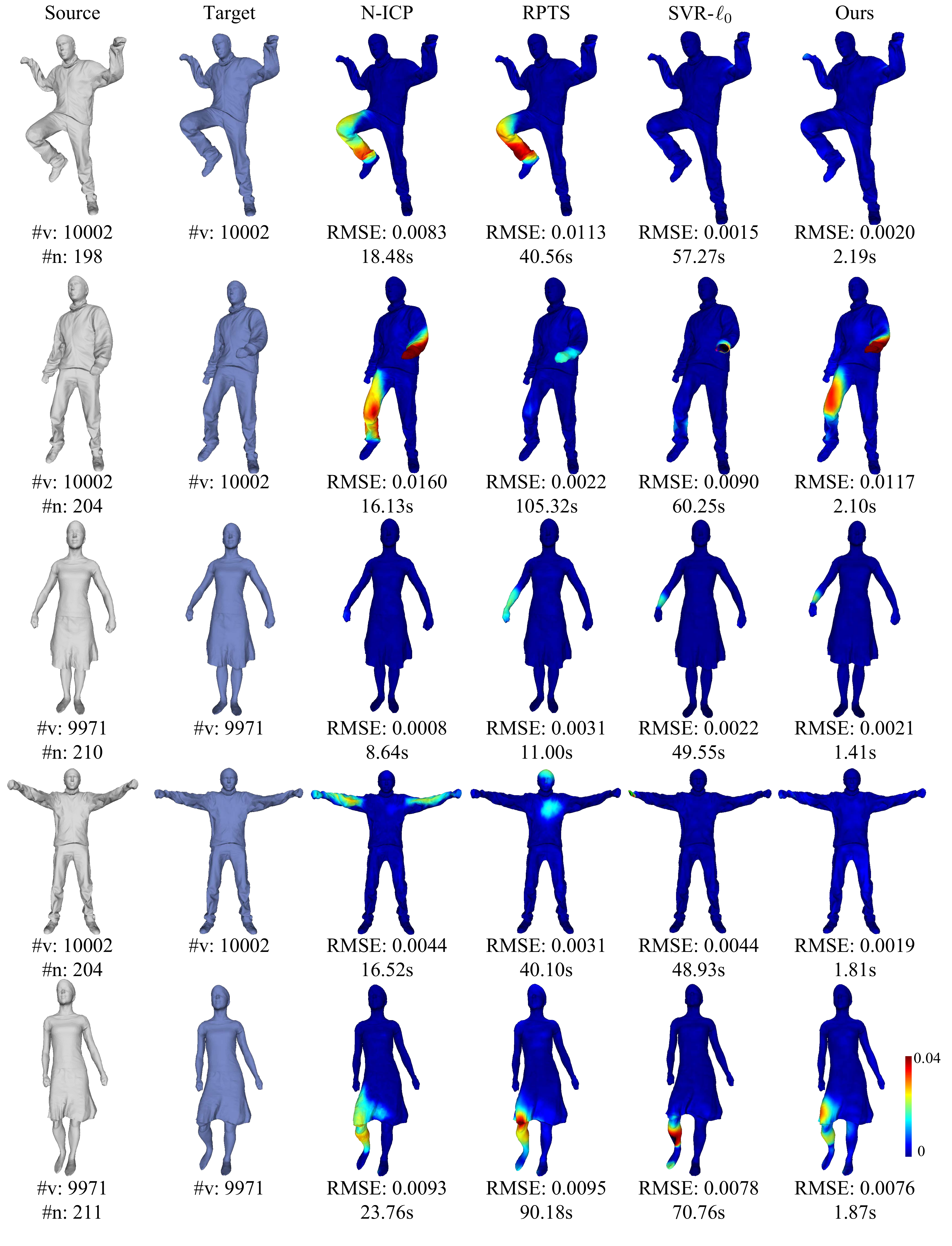}
    \caption{Comparison with N-ICP, RPTS and SVR-$\ell_0$ on ``crane'', ``march1'', ``samba'', ``squat1'' and ``swing'' datasets with small deformation. We set $\alpha = 10$ for N-ICP, $\alpha = 10$ and $\beta = 1$ for RPTS, $\alpha = 0.1$ and $\beta = 100$ for SVR-$\ell_0$, and $k_{\alpha} = 1, k_{\beta} = 10^3, \nua=30\overline{l}$ for our method in these examples. The RMSE and the color-coded registration errors are in the unit of meters.}
\label{fig:small}
\end{figure*}

\subsection*{Experiment on clean data}
\begin{figure*}[b]
    \centering
     \includegraphics[width=\textwidth]{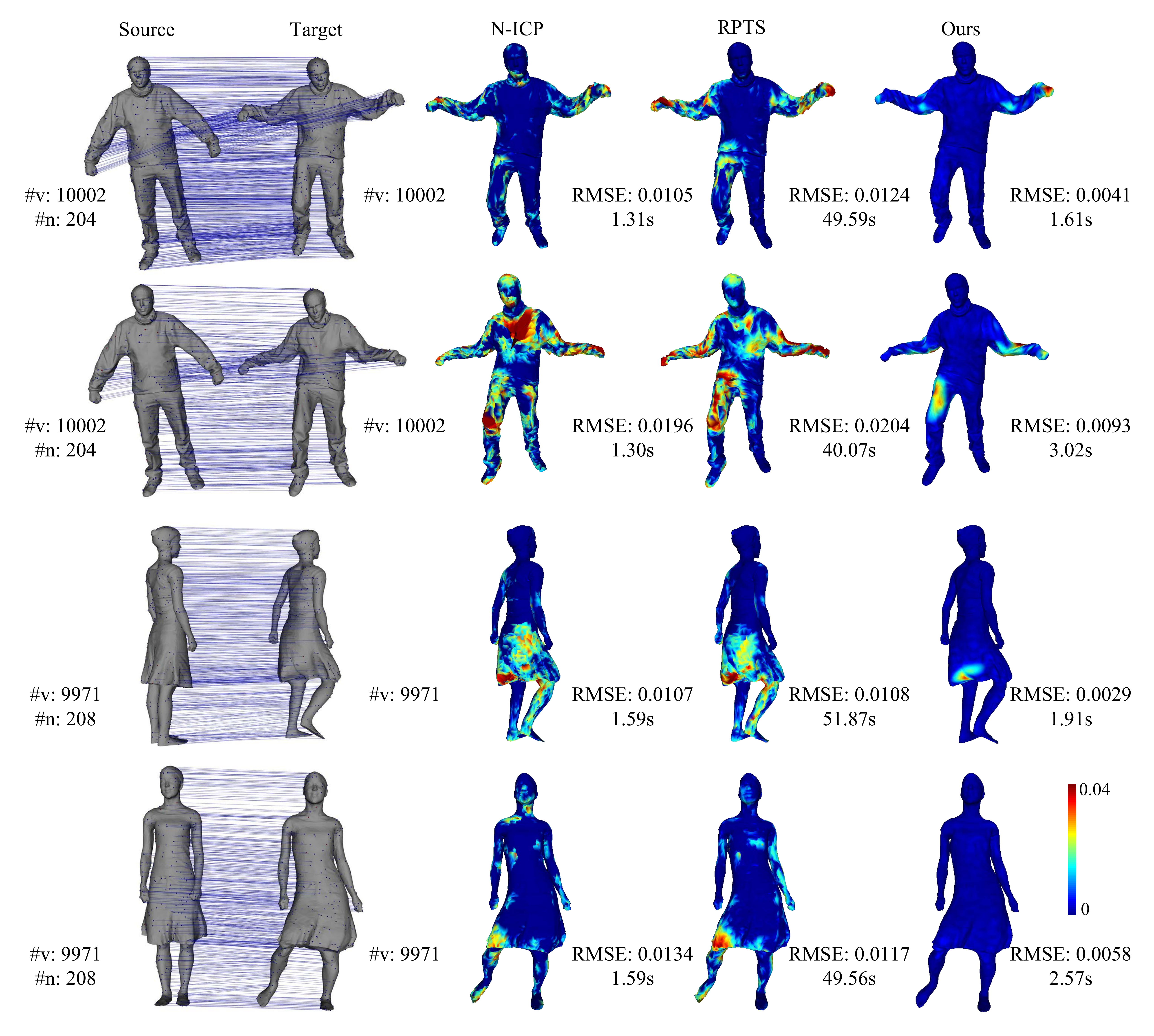}
    \caption{Comparison with N-ICP, RPTS and SVR-$\ell_0$ on ``crane'',and ``swing'' datasets with big deformation. We set $\alpha = 0.01$ for N-ICP, $\alpha = 0.01$ and $\beta = 1$ for RPTS, and $k_{\alpha} = 0.01$ and $k_{\beta} = 1$ for our method in these examples. The RMSE and the color-coded registration errors are in the unit of meters.}
\label{fig:big}
\end{figure*}

We show more results on five models ``crane'', ``march1'', ``samba'', ``squat1'' and ``swing'' in Human-motion datasets. For each model, we use the closest points to construct the correspondences for small deformation, and use the SHOT with diffusion pruning method for big deformation. For each method, we search the parameter setting for best performance. The results are shown in Fig.~\ref{fig:small} and Fig.~\ref{fig:big}, and we can see that our method is faster than other methods and achieves similar or better accuracy.

\subsection*{Experiment on partially overlapping data}
We show more results and comparisons on partially overlapping data from ``bouncing'' datasets. We choose $\alpha = 10$ for N-ICP, $\alpha = 1, \beta = 100$ for RPTS, $\alpha = 0.1, \beta = 100$ for SVR-$\ell_0$, and $k_{\alpha} = 1, k_{\beta} = 100, \nua^{\max}=30\overline{d}, \nur^{\max}=100\overline{l}$ for our method and $\theta=45^{\circ}$ for all method in these examples. The results are shown in Fig.~\ref{fig:partial}, and we can see that our methods is robust to partially overlapping data, and faster than other methods.
\begin{figure*}[h]
    \centering
     \includegraphics[width=0.9\textwidth]{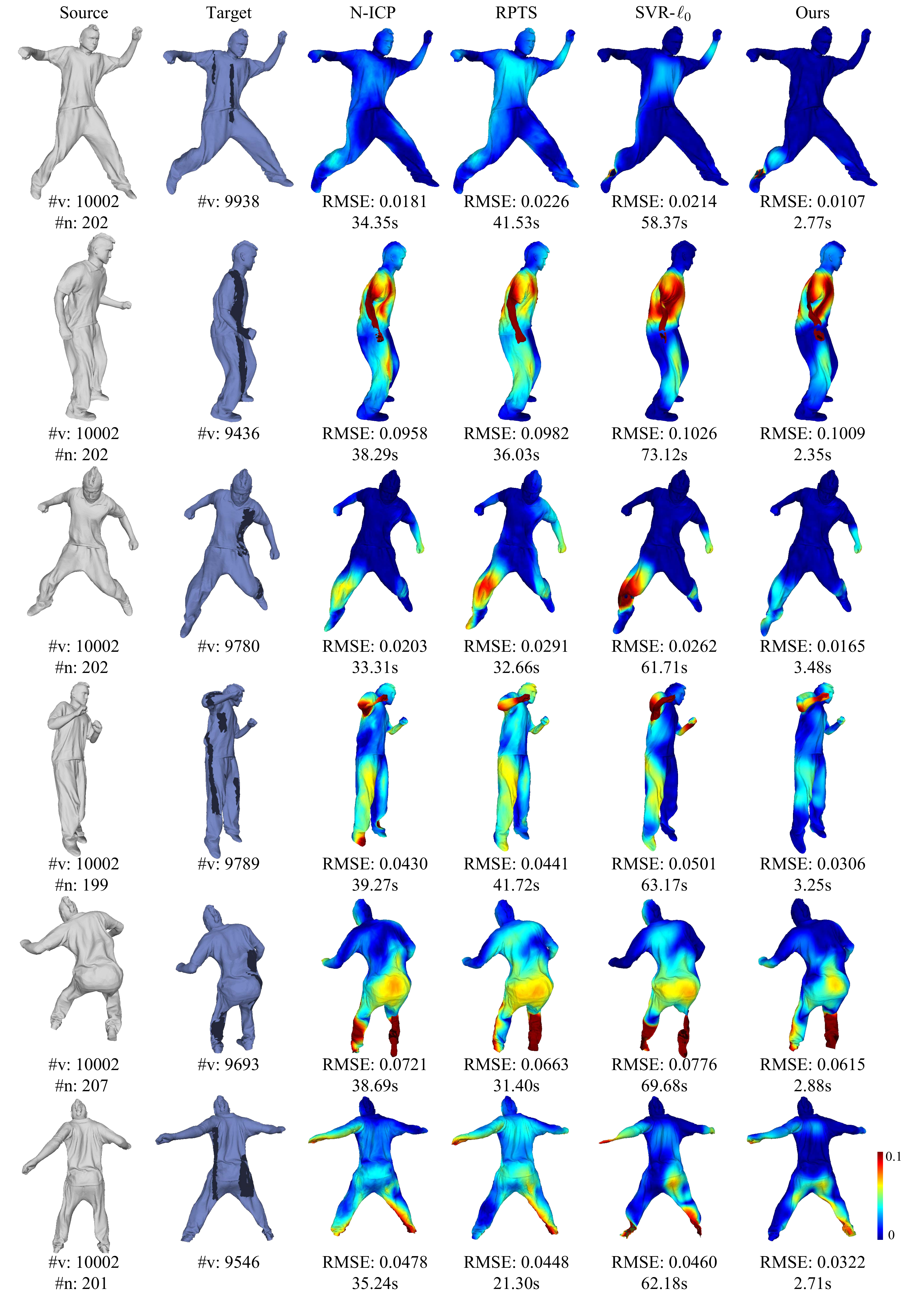}
    \caption{Comparison with N-ICP, RPTS and SVR-$\ell_0$ on ``bouncing''  datasets with partially overlapping data. The RMSE and the color-coded registration errors are in the unit of meters.}
    \label{fig:partial}
\end{figure*}

\end{document}